\documentclass{article} 
\PassOptionsToPackage{table}{xcolor}
\usepackage{iclr2025_conference,times}
\usepackage{amsmath,amsfonts,bm}
\usepackage{hyperref}
\usepackage{url}
\usepackage{xspace}
\usepackage{booktabs}
\usepackage{enumerate}
\usepackage{enumitem}
\usepackage{graphicx}
\usepackage{tcolorbox}
\usepackage{CJK}
\definecolor{darkgreen}{RGB}{0,100,0}

\newcommand{\algo}{\textsc{KgDG}\xspace}

\newcommand{\model}{\textsc{LawGPT}\xspace}
\newcommand{\Samp}{\textsc{KgGen}\xspace}
\newcommand{\Self}{\textsc{KgFix}\xspace}
\newcommand{\Veri}{\textsc{DaVer}\xspace}
\newcommand{\Mixt}{\textsc{MiTra}\xspace}

\title{\model: Knowledge-Guided Data Generation and Its Application to Legal LLM}

\author{
Zhi Zhou$^1$, Kun-Yang Yu$^{1,2}$, Shi-Yu Tian$^{1,2}$, Xiao-Wen Yang$^{1,2}$, Jiang-Xin Shi$^{1,2}$, \\
\textbf{Peng-Xiao Song$^1$, Yi-Xuan Jin$^1$, Lan-Zhe Guo$^{1,3,*}$, Yu-Feng Li$^{1,2,}$\thanks{Lan-Zhe Guo and Yu-Feng Li are corresponding authors.}}\\
$^1$National Key Laboratory for Novel Software Technology, Nanjing University, China\\
$^2$School of Artificial Intelligence, Nanjing University, China\\
$^3$School of Intelligence Science and Technology, Nanjing University, China\\
}

\iclrfinalcopy 
\begin{document}

\maketitle

\begin{abstract}
Large language models (LLMs), both proprietary and open-source, have demonstrated remarkable capabilities across various natural language processing tasks. However, they face significant limitations in legal reasoning tasks. Proprietary models introduce data privacy risks and high inference costs, while open-source models underperform due to insufficient legal domain training data. To address these limitations, we study data generation for legal reasoning to improve the legal reasoning performance of open-source LLMs with the help of proprietary LLMs. This is challenging due to the lack of legal knowledge in proprietary LLMs and the difficulty in verifying the generated data. We propose \algo, a knowledge-guided data generation framework for legal reasoning. Our framework enables leveraging legal knowledge to enhance generation diversity and introduces a refinement and verification process to ensure the quality of generated data. Moreover, we expand the generated dataset to further enhance the LLM reasoning capabilities. Using \algo, we create a synthetic legal reasoning dataset containing 50K high-quality examples. Our trained model \model outperforms existing legal-specific LLMs and achieves performance comparable to proprietary LLMs, demonstrating the effectiveness of \algo and \model. Our code and resources is publicly available at \url{https://github.com/LAMDASZ-ML/Knowledge-Guide-Data-Generation}.
\end{abstract}

\section{Introduction}

Large language models (LLMs)~\citep{GPT4,hugo23LLaMA2} have achieved remarkable success in various natural language processing tasks, including natural language understanding~\citep{dong19understanding}, reasoning~\citep{DBLP:conf/acl/0009C23}, and generation~\citep{yu22text}. 
Both proprietary and open-source LLMs exhibit strong generalization capabilities, enabling their application in diverse downstream scenarios, such as medicine~\citep{th23medicine}, finance~\citep{DBLP:journals/corr/abs-2306-06031}, education~\citep{DBLP:conf/bigdataconf/GanQWL23}. 
Recent studies~\citep{li2023LawBench,nguyen2023LawGPT} have demonstrated the preliminary effectiveness of existing general LLMs in legal reasoning tasks, including legal documents retrieval~\citep{chen2013textmining}, legal judgment prediction~\citep{luo2017legalpredict},  and legal question answering~\citep{zhong2020nlpbenefit}. 

Despite their preliminary success in legal reasoning applications, LLMs face significant practical limitations. Proprietary LLMs such as GPT-4~\citep{GPT4} and GPT-3.5 Turbo~\citep{GPT35T}, as well as extremely large open-source models like DeepSeek V3~\citep{deepseek24v3}, require API access, introducing substantial data privacy risks and high inference costs. Open-source LLMs like Qwen~\citep{yang24qwen25} and ChatGLM~\citep{du2022glm} show suboptimal performance due to training with insufficient legal data. These limitations create an opportunity to leverage proprietary LLMs for generating legal reasoning data to build open-source legal LLMs.

Previous studies have developed various data generation methods using proprietary LLMs for downstream reasoning tasks, which have been effective for mathematical reasoning~\citep{luo2025wizardmath}. 
These methods assume that the LLMs used for generation have sufficient knowledge about the downstream tasks and can generate diverse data through appropriate prompts~\citep{yu24metamath}. 
Moreover, for mathematical problems, their formal nature makes it straightforward to verify synthetic data~\citep{li2024neurosymbolic} and eliminate incorrect data caused by hallucination issue. 
However, legal reasoning presents unique challenges: 
LLMs for generation lack specific legal knowledge, which limits the diversity of synthetic data. 
Additionally, the informal and complex nature of legal reasoning makes it difficult to formalize and verify the generated data.

To address these challenges, we propose \algo, a \emph{\underline{K}nowledge-\underline{g}uided \underline{D}ata \underline{G}eneration} framework for legal reasoning tasks. 
Our framework consists of three key components: (1) \emph{\underline{K}nowledge-\underline{G}uide \underline{Gen}eration} (\Samp), which leverages a legal knowledge base $\mathcal{K}$ to generate diverse data; 
(2) \emph{\underline{K}nowledge-\underline{G}uide \underline{Fix}er} (\Self), which corrects incorrect references and reasoning paths; 
and (3) \emph{\underline{Da}ta \underline{Ver}ifier} (\Veri), which filters out uncorrectable data to ensure generation quality. To further enhance the reasoning capabilities of trained LLMs, we propose a \emph{\underline{Mi}xture \underline{Tra}ining} (\Mixt) strategy that expands the generated dataset.
Using \algo, we create a synthetic legal reasoning dataset containing 50K high-quality examples. 
Our trained model \model outperforms existing legal-specific LLMs and achieves performance comparable to proprietary LLMs, demonstrating the effectiveness of both \algo and \model. Our contributions can be summarized as follows:
\begin{enumerate}[label=(\alph*),itemsep=0pt,topsep=0pt,parsep=0pt]
    \item We propose \algo, a knowledge-guided data generation framework that enables the creation of high-quality and diverse datasets for legal reasoning tasks, addressing the challenges of limited generation diversity and difficulty in verifying generated data. 
    \item We create a large-scale synthetic dataset using \algo and train \model with different model scales. The dataset and models will be publicly available to facilitate future research. 
    \item Extensive experiments demonstrate \model outperforms state-of-the-art legal-specific LLMs and achieves comparable performance to proprietary LLMs in legal reasoning. 
\end{enumerate}

\section{Methodology}

In this section, we introduce \algo, an LLM-based data generation framework, building data to improve the legal reasoning performance of open-source LLMs. However, the following two challenges make it difficult for exising LLMs to generate data for legal reasoning:
\begin{enumerate}[label=(\alph*),itemsep=2pt,topsep=0pt,parsep=0pt]
    \item LLMs for data generationlack legal knowledge, which limits the diversity 
     of synthetic data.
    \item Legal synthetic data is difficult to formalize and verify, making it challenging to detect and eliminate hallucinations in the generation process.
\end{enumerate}

\begin{figure}[t]
   \begin{center}
      \includegraphics[width=0.95\textwidth]{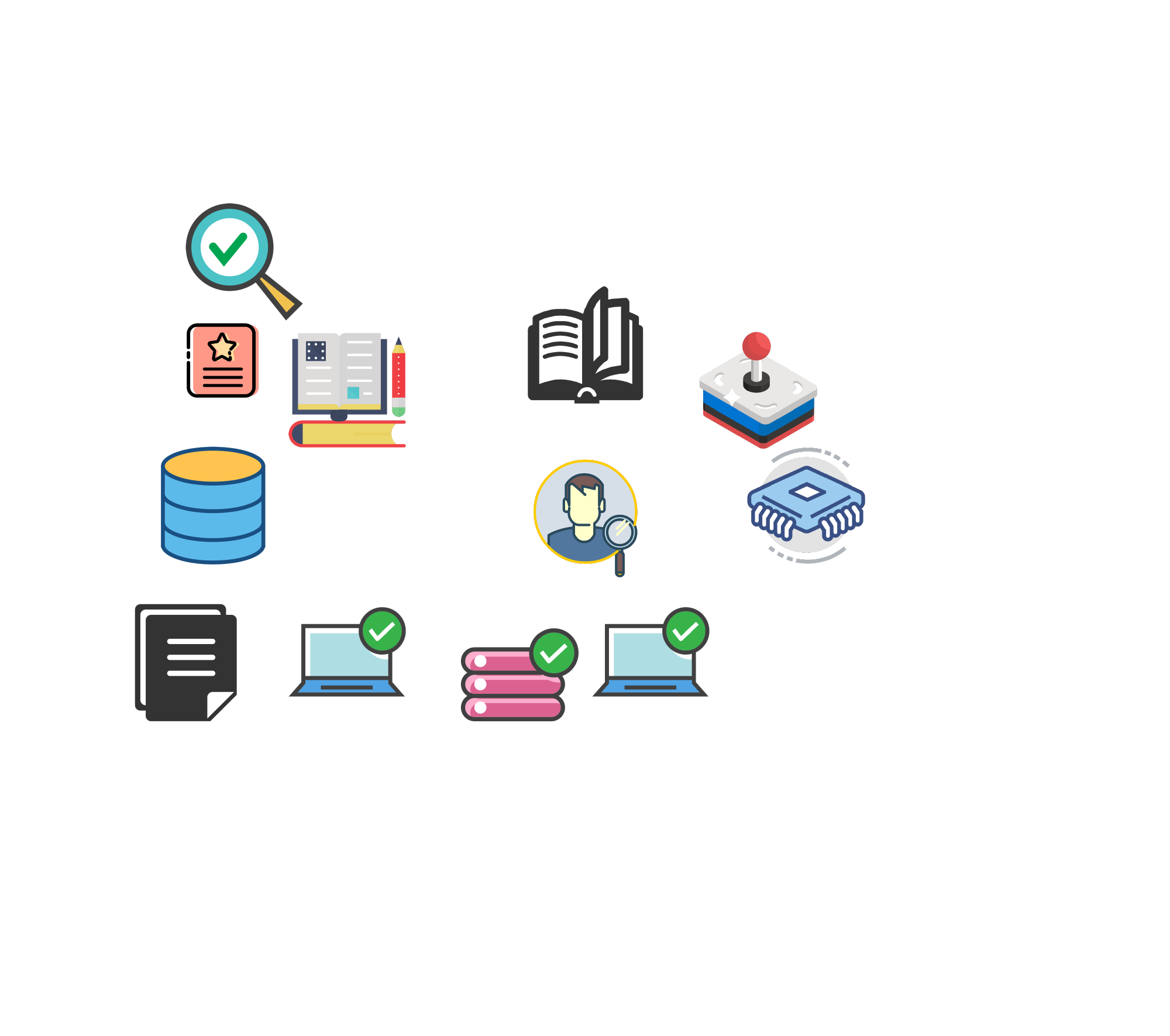}
   \end{center}
   \caption{Illustration of \algo, a knowledge-guided data generation framework. }
   \label{fig: data-generation-framework}
\end{figure}

We design \emph{\underline{K}nowledge-\underline{G}uided \underline{Gen}eration} (\Samp) to address the first challenge by introducing legal documents as knowledge base. Then, \emph{\underline{K}nowledge-\underline{G}uided \underline{Fix}er} (\Self) and \emph{\underline{Da}ta \underline{Ver}ifier} (\Veri) addressing the second challenge by refining correctable errors and removing uncorrectable data. 
To further improve model reasoning performance, we implement a \emph{\underline{Mi}xture \underline{Tra}ining} (\Mixt) to teach open-source LLMs to reason step-by-step while keeping the capability to directly generate answers efficiently.
Overall illustration is shown in Figure~\ref{fig: data-generation-framework} and each module is detailed below.

\subsection{Knowledge-Guided Generation (\Samp)}

Existing studies~\citep{li2024neurosymbolic} demonstrate that data generation methods based on LLMs have strong potential for building high-quality training data. However, for tasks that require specific domain knowledge, such as legal reasoning, LLMs may fail to build high-quality data due to their lack of domain knowledge, leading to insufficient diversity in synthetic data. 
To address this challenge, we design \Samp by introducing a knowledge base $\mathcal{K}$ to compensate for the lack of legal knowledge inherent in LLMs. 
This enables us to generate diverse synthetic data by leveraging legal knowledge sampling on the knowledge base $\mathcal{K}$. Specifically, for legal reasoning task, \Samp consists of two components: \emph{Knowledge-Aware Sampler} and \emph{Knowledge-Guided Writer}. The \emph{Knowledge-Aware Sampler} employs sampling strategies to enhance the diversity of synthetic data, while the \emph{Knowledge-Guided Writer} leverages LLMs to extract core information and generate question-answer pairs.

The \emph{Knowledge-Aware Sampler} takes two inputs: a knowledge base $\mathcal{K}$ containing legal documents and a seed problem set $\mathcal{E}$ providing format examples for legal reasoning tasks.
The sampling process is controlled by a strategy $\pi(\mathbf{k}, \mathbf{e}|\mathcal{D}_{\mathrm{Gen}})$ that samples from $\mathcal{K}$ and $\mathcal{E}$ conditioned on the current generated dataset $\mathcal{D}_{\mathrm{Gen}}$, where $\mathbf{k} \in \mathcal{K}$ represents a sampled legal document and $\mathbf{e} \in \mathcal{E}$ represents a sampled seed problem. We implement $\pi$ as a two-step sampling strategy: (1) LLM selects specific types of legal knowledge from $\mathcal{K}$ based on the sampled example problem $\mathbf{e}$ to ensure consistency between the example and knowledge; (2) Monte Carlo sampling ensures diverse and balanced synthetic data across all problem types and their corresponding legal knowledge domains.

The \emph{Knowledge-Guided Writer} $\mathrm{LLM}_{\mathrm{W}}$ takes the sampled legal document $\mathbf{k}$ and example problem $\mathbf{e}$ as inputs and generates the unverified draft data $\tilde{\mathbf{x}}$ containing question $\tilde{\mathbf{q}}$, answer $\tilde{\mathbf{a}}$, reasoning path $\tilde{\mathbf{p}}$, and references $\tilde{\mathbf{r}}$:
\begin{equation}
   \tilde{\mathbf{x}} = (\tilde{\mathbf{q}}, \tilde{\mathbf{a}}, \tilde{\mathbf{r}}, \tilde{\mathbf{p}}) = \mathrm{LLM}_{\mathrm{W}}(\mathbf{k}, \mathbf{e})
\end{equation}

\subsection{Knowledge-Guide Fixer (\Self) and Data Verifier (\Veri)}

The unverified draft data $\tilde{\mathbf{x}} = (\tilde{\mathbf{q}}, \tilde{\mathbf{a}}, \tilde{\mathbf{r}}, \tilde{\mathbf{p}})$ contains potential errors in all components due to the hallucination problems of LLMs. 
To address this issue, we introduce \Self to fix correctable errors in the reasoning path $\tilde{\mathbf{p}}$ and references $\tilde{\mathbf{r}}$, and \Veri to filter out uncorrectable data.

\Self consists of two components: \emph{Reference Modifier} and \emph{Reasoning Corrector}. 
The \emph{Reference Modifier} validates and corrects legal references using LLMs or the knowledge base, generating a corrected reference $\hat{r} = \mathrm{Fixer}_{\mathrm{M}}(\tilde{r})$. 
The \emph{Reasoning Corrector} examines the reasoning path for logical and computational errors using LLMs or tools, producing a corrected path $\hat{p} = \mathrm{Fixer}_{\mathrm{C}}(\tilde{p})$.  

While \Self ensures the correctness of reference $\hat{r}$ and reasoning path $\hat{p}$, it cannot guarantee their relevance to the generated question-answer pair $(\tilde{q}, \tilde{a})$. 
Therefore, we implement \Veri to validate whether the answer $\tilde{a}$ can be derived from the question $\tilde{q}$ using the corrected references $\hat{r}$ and reasoning path $\hat{p}$. If the validation succeeds, we mark the question-answer pair as valid (denoted as $\hat{q}$ and $\hat{a}$). The verified data $\hat{x} = (\hat{q}, \hat{a}, \hat{r}, \hat{p})$ is then added to the synthetic dataset $\mathcal{D}_{\mathrm{Gen}}$. This process continues until $|\mathcal{D}_{\mathrm{Gen}}|$ meets the required data volume.

\subsection{Mixture Training (\Mixt)}

To further enhance the reasoning performance of the trained LLM, we implement \Mixt to generate two types of training data using $\mathcal{D}_{\mathrm{Gen}}$: 
(1) standard question-answer pairs and 
(2) question-answer pairs with explicit reasoning paths. 
The standard pairs enable efficient direct responses, while the pairs with reasoning paths teach the model step-by-step reasoning.

Specifically, we design two prompt templates: $\mathrm{T}_s(\hat{\mathbf{q}}, \hat{\mathbf{a}})$ for standard pairs and $\mathrm{T}_r(\hat{\mathbf{q}}, \hat{\mathbf{a}}, \hat{\mathbf{r}}, \hat{\mathbf{p}})$ for pairs with reasoning paths. 
Here, $\mathrm{T}_s$ generates training instances using only questions and answers, while $\mathrm{T}_r$ incorporates additional reasoning steps seperated by a thinking tag in the responses. 
Example problems for both types of training data are provided in Appendix~\ref{app:examples}.
The final training dataset is constructed by combining both types:
\begin{equation}
   \mathcal{D}_{\mathrm{Train}} = \left\{ \mathrm{T}_s(\hat{\mathbf{q}}, \hat{\mathbf{a}}) \right\} \cup \left\{ \mathrm{T}_r(\hat{\mathbf{q}}, \hat{\mathbf{a}}, \hat{\mathbf{r}}, \hat{\mathbf{p}}) \right\}, \quad (\hat{\mathbf{q}}, \hat{\mathbf{a}}, \hat{\mathbf{r}}, \hat{\mathbf{p}}) \sim \mathcal{D}_{\mathrm{Gen}}
\end{equation}

\section{Experiments}
In this section, we compare the performance of \model against base models, legal-specific LLMs, and general LLMs to demonstrate the effectiveness of \algo framework and the trained \model. 

\subsection{Experimental Settings}

\paragraph{Evaluation Protocol.} To evaluate the legal reasoning performance of each model, we adopt four legal reasoning tasks: Scene-based Article Prediction (Task \#1)~\citep{LAWGPT-zh}, Prison Term Prediction without Article (Task \#2), Prison Term Prediction with Article (Task \#3)~\citep{cail2018}, and Criminal Damages Calculation (Task \#4)~\footnote{\url{https://laic.cjbdi.com/}}. 
Task \#1 is evaluated using the ROUGE-L score to compare the legal article prediction with the ground truth. Tasks \#2 and \#3 are evaluated using Normalized log-distance to compare the predicted prison term. Task \#4 is evaluated using accuracy to determine whether the predicted damages match the ground truth. The implementation of our evluation is based on the LawBench~\citep{li2023LawBench}.

\paragraph{Comparison Models.} We compare two types of models: (1) General proprietary LLMs, including GPT-4~\citep{GPT4},  GPT-3.5 Turbo~\citep{GPT35T}, and DeepSeek V3~\citep{deepseek24v3}; (2) Legal-specific LLMs, including Lexilaw~\citep{li24lexilaw}, LaywerLLaMA~\citep{huang24lawyerllama}, HanFei~\citep{zhang2023HanFei}, ChatLaw~\citep{cui23chatlaw}, FuziMingcha~\citep{deng2023fuzi}, and WisdomInterrogatory~\citep{wu2024WisdomInterrogatory}.

\paragraph{Dataset Construction.} 
We implement the \algo framework using the DeepSeek V3 model~\citep{deepseek24v3}, based on a legal knowledge base and a constructed seed problem set. Specifically, to construct the legal knowledge base, we manually collect 186,197 high-quality criminal legal documents and 152,452 civil legal documents. Each document includes judgment facts, reasons, results, and relevant laws. This knowledge base supports the generation of diverse and synthetic problems for legal reasoning, as well as the verification and correction of generated reasoning paths and answers. 
For seed problems, we manually construct ten problems for each task as examples to guide the \algo to generate legal problems in the desired format. These seed problems are solely for demonstration and are not used for training. 
\algo generates 25K legal problems with verified answers. 
\Mixt expands each problem into two: one with direct answers and one with answers accompanied by detailed reasoning steps, resulting in a total of 50K training examples.
The detailed implementation of \algo and generation process is provided in Appendix~\ref{app:implementation}.

\paragraph{Model Training.} We adopts the LLaMA-Factory~\citep{zheng24llamafactory} to fine-tune the series of Qwen-2.5 models~\citep{yang24qwen25}, including 0.5B, 1.5B, and 3B versions. The training epochs are set to 3 and learning rate is set to 1e-5 with a cosine learning rate scheduler. Our training process is conducted on a Linux server with 4 NVIDIA A800 GPUs.

\subsection{Empirical Results}

In this section, we conduct experiments to compare the performance of \model with base models, general LLMs, and legal-specific LLMs to demonstrate the effectiveness of our \algo framework as well as the trained legal LLM \model. 

\begin{figure}[t]
   \begin{center}
      \includegraphics[width=0.9\textwidth]{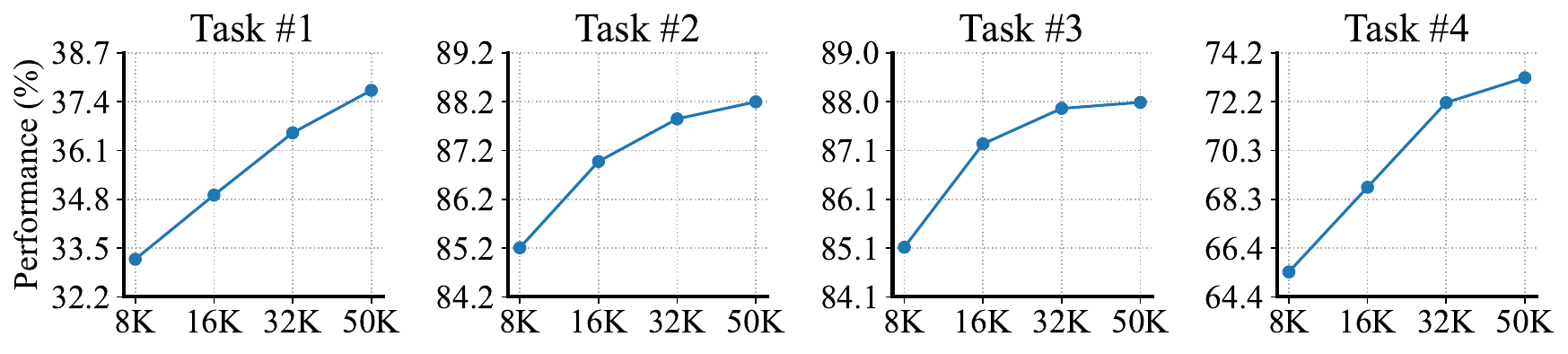}
   \end{center}
   \vskip -1em
   \caption{Scalability analysis of the \algo framework. The performance on all tasks improves as the amount of generated training data increases.}
   \label{fig: data-scalability}
\end{figure}

\paragraph{Effectiveness of \algo.} To evaluate the effectiveness of our proposed \algo data generation framework, we fine-tune Qwen-2.5 models of different scales using our generated 50K data. The results in Table~\ref{tab:scale} demonstrate that out fine-tuned model consistently outperforms the base models across all scales. This indicates that \algo generates high-quality legal data that effectively improves the reasoning capabilities of base models regardless of their size. 
Moreover, we analyze the scalability of \algo in Figure~\ref{fig: data-scalability}. The experimental results demonstrate that the performance of trained LLMs consistently improves across all tasks as the volume of generated training data increases, indicating the strong potential of \algo for developing more capable legal LLMs.

\paragraph{Effectiveness of \model.} We evaluate \model against both general and legal-specific LLMs. For general LLMs, we include two proprietary models (GPT-4 and GPT-3.5 Turbo) and one large-scale open-source model (DeepSeek V3). We also compare against seven legal-specific LLMs of various sizes. As shown in Table~\ref{tab:performance}, \model outperforms all existing legal-specific LLMs despite its smaller scale. Furthermore, \model surpasses GPT-4 and GPT-3.5 Turbo while achieving performance comparable to DeepSeek V3 on multiple tasks. These results demonstrate both the value of specialized legal LLMs and the effectiveness of our \algo framework.

\begin{table}[!t]
   \caption{Performance comparison between \model and Qwen-2.5 base models of different model scales. \model consistently outperforms Qwen-2.5 across all model sizes and tasks, showing the effectiveness of our \algo framework.}
   \label{tab:scale}
   \begin{center}
   \begin{tabular}{lr|rrrrr}
   \bottomrule
   \toprule
   Models & \#Parameters & Task \#1 & Task \#2 & Task \#3 & Task \#4 & Average\\
   \midrule
   Qwen-2.5 & 0.5B & 27.9 & 81.2 & 80.1 & 45.0 & 58.6 \\
   \model & 0.5B & 33.1 & 86.8 & 86.6 & 62.0 & 67.1 \\ \hline
   \rowcolor{gray!20} \multicolumn{2}{c|}{$\Delta$ Performance} & \textcolor{darkgreen}{$\uparrow$ 5.2} & \textcolor{darkgreen}{$\uparrow$ 5.6} & \textcolor{darkgreen}{$\uparrow$ 6.5} & \textcolor{darkgreen}{$\uparrow$ 14.0} & \textcolor{darkgreen}{$\uparrow$ 9.5} \\
   \midrule
   Qwen-2.5 & 1.5B & 29.9 & 82.4 & 82.3 & 49.0 & 60.9 \\
   \model   & 1.5B & 35.7 & 87.4 & 87.3 & 68.0 & 69.6 \\ \hline
   \rowcolor{gray!20} \multicolumn{2}{c|}{$\Delta$ Performance} & \textcolor{darkgreen}{$\uparrow$ 5.8} & \textcolor{darkgreen}{$\uparrow$ 5.0} & \textcolor{darkgreen}{$\uparrow$ 5.0} & \textcolor{darkgreen}{$\uparrow$ 19.0} & \textcolor{darkgreen}{$\uparrow$ 8.7} \\
   \midrule
   Qwen-2.5 & 3.0B & 28.7 & 81.7 & 79.9 & 56.0 & 61.6 \\
   \model   & 3.0B & 37.7 & 88.2 & 88.0 & 73.2 & 71.8 \\ \hline
   \rowcolor{gray!20} \multicolumn{2}{c|}{$\Delta$ Performance} & \textcolor{darkgreen}{$\uparrow$ 9.0} & \textcolor{darkgreen}{$\uparrow$ 6.5} & \textcolor{darkgreen}{$\uparrow$ 8.1} & \textcolor{darkgreen}{$\uparrow$ 17.2} & \textcolor{darkgreen}{$\uparrow$ 10.2} \\
   \bottomrule
   \toprule
   \end{tabular}
   \end{center}
   \vskip -1em
\end{table}

\begin{table}[!t]
   \caption{Performance comparison between \model and general LLMs and legal-specific LLMs. The results show that \model outperforms exisings legal-specific LLMs. Moreover, \model can achieves similar performance to general LLMs even with a significantly smaller scale.
   The best performance is highlighted in bold and the second best is underlined among legal-specific LLMs.}
   \label{tab:performance}
   \vskip -1em
   \begin{center}
\begin{tabular}{lr|ccccc}
   \bottomrule
   \toprule
   Models & \#Parameters & Task \#1 & Task \#2 & Task \#3 & Task \#4 & Average\\
   \midrule
   \multicolumn{7}{c}{\bf General LLMs} \\
   \midrule
   \rowcolor{gray!20} Deepseek V3   & 671B & 38.1 & 87.5 & 86.8 & 84.4 & 74.2 \\
   \rowcolor{gray!20} GPT-4         & - & 27.5 & 82.6 & 81.9 & 77.6 & 67.4 \\
   \rowcolor{gray!20} GPT-3.5 Turbo & - & 31.3 & 78.7 & 76.8 & 61.2 & 62.0 \\
   \midrule
   \multicolumn{7}{c}{\bf Legal-Specific LLMs} \\
   \midrule
   Lexilaw & 7B & \underline{35.8} & 78.1 & 74.9 & 35.8 & 56.1 \\
   HanFei & 7B & 33.6 & 73.1 & 69.6 & 39.4 & 53.9 \\
   FuziMingcha & 7B & 22.2 & 77.2 & 75.5 & 47.2 & 55.5 \\
   WisdomInterrogatory & 7B & 32.0 & 80.4 & 81.1 & 17.4 & 52.7 \\
   LaywerLLaMA & 13B & 25.9 & 74.2 & 75.5 & 39.2 & 53.7 \\
   ChatLaw & 13B & 31.6 & 76.2 & 73.6 & 41.4 & 55.7 \\
   ChatLaw & 33B & 26.0 & 67.0 & 53.6 & 41.6 & 47.1 \\
   \midrule
   \model & 0.5B & 33.1 & 86.8 & 86.6 & 62.0 & 67.1 \\
   \model & 1.5B & 35.7 & \underline{87.4} & \underline{87.3} & \underline{68.0} & \underline{69.6} \\
   \model & 3B & \textbf{37.7} & \textbf{88.2} & \textbf{88.0} & \textbf{73.2} & \textbf{71.8} \\
   \bottomrule
   \toprule
   \end{tabular}
\end{center}
\vskip -1em
\end{table}

\subsection{Ablation Study}

We conduct an ablation study using a 4K subset of the training data to evaluate the effectiveness of each component in our \algo framework. The results are shown in Table~\ref{tab:ablation}. The model achieves its best average performance only when all four modules are integrated. For Task \#2 and \#3, we observe that the \Veri module introduces a slight performance degradation when handling complex prison term prediction tasks, indicating potential room for improvement in this module. Nevertheless, the integration of all four modules still yields the best overall performance, demonstrating the value of each component in our \algo framework.

\begin{table}[!t]
   \caption{Ablation study. We conduct experiments on the Qwen-2.5-3B model using a 4K subset of generated data. Our four proposed modules are added sequentially to assess their effectiveness. The results show that the best average performance is achieved when all four modules are integrated.}
   \label{tab:ablation}
   \vskip -1em
   \begin{center}
   \resizebox{1.0\textwidth}{!}{
      \begin{tabular}{cccc|cccc|c}
         \bottomrule
         \toprule
         \Samp & \Self & \Veri & \Mixt & Task \#1 & Task \#2 & Task \#3 & Task \#4 & Average \\
         \midrule
         & & & & 28.7 & 81.7 & 79.9 & 56.0 & 61.6 \\
         \checkmark & & & & 30.9 & 84.9 & 84.7 & 59.8 & 65.1 \\
         \checkmark & \checkmark & & & 31.1 & \textbf{85.6} & \textbf{85.4} & 60.4 & 65.6 \\
         \checkmark & \checkmark & \checkmark & & \underline{31.5} & 85.1 & 84.8 & \underline{65.0} & \underline{66.6} \\
         \checkmark & \checkmark & \checkmark & \checkmark & \textbf{33.2} & \underline{85.2} & \underline{85.1} & \textbf{65.4} & \textbf{67.2} \\
         \bottomrule
         \toprule
      \end{tabular}}
   \end{center}
   \vskip -1em
\end{table}

\section{Conclusion}

In this paper, we study data generation for legal reasoning to improve the performance of open-source legal LLMs with the help of proprietary LLMs. 
To address the challenges of limited diversity in synthetic legal data and the difficulty of data verification, we propose \algo, a knowledge-guided data generation framework. 
Our framework consists of three key components that leverage legal knowledge to enhance generation diversity and ensure data quality through refinement and verification processes. Additionally, we develop \Mixt to expand the generated dataset and further enhance LLM reasoning capabilities. Both \algo and \model are validated by extensive experiments on multiple legal reasoning tasks. \model achieves comparable performance to proprietary LLMs while being significantly smaller in scale.

\textbf{Limitations and Future Work.} This paper gives a preliminary study on the data generation for legal LLMs and we only make a simple attempt to build each component in the \algo framework, which is mainly relies on prompting LLMs, and could be further improved by incorporating more sophisticated techniques. 
Moreover, our current study is limited to generating 50K training examples and training models with less than 3B parameters. While this scale is sufficient to validate the effectiveness of \algo and \model, future work could explore the upper bound of our framework by scaling up both the size of synthetic dataset and trained model size.

\bibliography{ref}
\bibliographystyle{iclr2025_conference}

\newpage
\appendix

\newcounter{promptnumber}
\newtcolorbox{promptbox}[1][]{
    colback=gray!5,
    colframe=gray!50,
    coltitle=black,
    boxrule=0.5pt,
    arc=0mm,
    left=5pt,
    right=5pt,
    top=5pt,
    bottom=5pt,
    title={Prompt~\thepromptnumber},
    #1
}

\section{Implementation Details for \algo}
\label{app:implementation}

We implement the \algo framework based on DeepSeek V3 model~\citep{deepseek24v3} and a knowledge based with 186,197 high-quality criminal legal documents and 152,452 civil legal documents. In our implementation, we call API of DeepSeek V3 model in parallel with a batch size of 16 and the generation process repeats until the number of generated data reaches 25K.
The specific implementation details are as follows. 

\paragraph{\Samp.} We first use the Prompt for Generation of \Samp to select which type of legal document should be sampled to generate similar types of reasoning problems based on the example.

\begin{CJK*}{UTF8}{gkai}
\begin{promptbox}[title=Prompt for Sampling of \Samp]
给你一个 JSON 格式的法律领域的问题及其答案。其中，instruction\ 字段指导如何回答问题，question\ 字段中包含问题，answer\ 字段中包含答案。\\
\{JSON\}\\
现在请你根据法律文书数据生成类似的问题，请问你需要什么类型的文书数据。可以选择的类型有：刑事法律文书、民事法律文书。请你选择一项并以\ JSON\ 格式在\ type\ 字段中返回。
\end{promptbox}
\end{CJK*}
Here, the example problem is provided in JSON format in `\{JSON\}'. The \emph{Knowledge-Aware Sampler} first determines the appropriate legal document type based on the example problem. Then, it randomly samples a document from the knowledge base of that type and generates a new problem-answer pair, complete with extracted references and reasoning paths.

\begin{CJK*}{UTF8}{gkai}
\begin{promptbox}[title=Prompt for Generation of \Samp]
给你一个 JSON 格式的法律领域的问题及其答案。其中，instruction\ 字段指导如何回答问题，question\ 字段中包含问题，answer\ 字段中包含答案。\\
\{JSON\}\\
请你以如下法律文书的内容为原型，按照相同的\ JSON\ 格式和问题形式，在\ instruction\ 不变的情况下，编造一个新问题与对应的答案。\\
请增加一个\ reasoning\ 字段，此字段是一个字符串，表示得出答案的推理过程。\\
请增加一个\ reference\ 字段，此字段是一个字典，Key 为推理过程中涉及的法律法条，Value 表示法律法条的具体内容。\\
请适当改写法律文书的内容，不要包含与答案无关的内容，不要直接复述法律文书的内容。\\
请修改问题与答案中的姓名、企业名称、地点等涉及隐私的内容。\\
answer\ 字段的内容应该完全按照\ instruction\ 中的对答案的格式要求给出。\\
\{DOCS\}
\end{promptbox}
\end{CJK*}
Here, the example problem is provided in JSON format in `\{JSON\}' and the sampled legal document is provided in `\{DOCS\}'.

\paragraph{\Self.} We first use the Prompt for Reference Modifier and Reasoning Corrector to correct the references and reasoning paths for each draft data.

\begin{CJK*}{UTF8}{gkai}
\begin{promptbox}[title=Prompt for Reference Modifier]
给你一个包含若干法条的 JSON 字典，此字段是一个字典，Key\ 为推理过程中涉及的法律法条，Value\ 表示法律法条的具体内容。\\
\{JSON\}\\
法条的内容可能存在问题，请你将\ Value\ 修正为\ Key\ 对应的正确法条内容，并以\ JSON\ 格式返回，不要附加其他内容或说明。
\end{promptbox}
\end{CJK*}

\begin{CJK*}{UTF8}{gkai}
\begin{promptbox}[title=Prompt for Reasoning Corrector]
给你一个 JSON 格式的法律领域的问题及其答案。其中，instruction\ 字段指导如何回答问题，question\ 字段中包含问题，answer\ 字段中包含答案，reference\ 字段中包含法律法条的内容，reasoning\ 包含推理过程。\\
\{JSON\}\\
当前问题的推理过程与答案可能存在问题，请根据问题内容、法律法条内容，改进当前的推理过程与答案。\\
如果此问题的推理过程与答案无需改进，请直接输出原始\ JSON\ 格式内容，否则请修改\ reasoning\ 字段和\ answer\ 字段的内容后，直接输出\ JSON\ 格式内容。不要附加其他内容或说明。
\end{promptbox}
Here, the draft data is provided in JSON format in `\{JSON\}'. 
\end{CJK*}

\paragraph{\Veri.} We first use the Prompt for Verification to verify the correctness of the generated question-answer pair as well as the consistency between the reasoning, reference and the answer.

\begin{CJK*}{UTF8}{gkai}
\begin{promptbox}[title=Prompt for Verification]
给你一个 JSON 格式的法律领域的问题及其答案。其中，instruction\ 字段指导如何回答问题，question\ 字段中包含问题，answer\ 字段中包含答案，reference\ 字段中包含法律法条的内容，reasoning\ 包含推理过程。
\{JSON\}\\
请你判断数据中的推理过程与答案是否正确，请以\ JSON\ 格式返回你的判断结果。JSON格式数据中包含一个\ verify\ 字段，取值为正确或错误，也包含一个\ message\ 字段，表示你判断的理由。   
\end{promptbox}
\end{CJK*}
Here, the draft data to be verified is provided in JSON format in `\{JSON\}'.

\newtcolorbox{problembox}[1][]{
    colback=blue!5,
    colframe=blue!50,
    coltitle=black,
    boxrule=0.5pt,
    arc=0mm,
    left=5pt,
    right=5pt,
    top=5pt,
    bottom=5pt,
    title={Problem~\thepromptnumber},
    #1
}

\section{Examples of Synthetic Data}
\label{app:examples}
\begin{CJK*}{UTF8}{gkai}
   \begin{problembox}[title=Standard Question-Answer Pair]
\textbf{Problem:}\\
请你仔细计算文书中涉及的犯罪总金额。无需给出计算过程，只需要给出最终金额，将答案写在[金额]与$<$eoa$>$之间，例如[金额]2000元$<$eoa$>$。文书:经审理查明，2018年5月12日晚20时许，被告人张某某酒后驾驶摩托车在市区XX路行驶时，与行人李某发生碰撞，造成李某重伤。事故发生后，张某某逃离现场。经鉴定，李某的医疗费用为15000元，后续治疗费用预计为5000元，误工费为3000元，护理费为2000元。张某某在案发后支付了李某的医疗费用15000元。\\
\textbf{Answer:} \newline [金额]25000元$<$eoa$>$
   \end{problembox}
\end{CJK*}

\begin{CJK*}{UTF8}{gkai}
   \begin{problembox}[title=Question-Answer Pair with Reasoning Path]
\textbf{Problem:}\\
请你给出回复的时候，在$<$DTK$>$标签前给出你的思考过程后再作答。请你仔细计算文书中涉及的犯罪总金额。无需给出计算过程，只需要给出最终金额，将答案写在[金额]与$<$eoa$>$之间，例如[金额]2000元$<$eoa$>$。文书:经审理查明，2018年5月12日晚20时许，被告人张某某酒后驾驶摩托车在市区XX路行驶时，与行人李某发生碰撞，造成李某重伤。事故发生后，张某某逃离现场。经鉴定，李某的医疗费用为15000元，后续治疗费用预计为5000元，误工费为3000元，护理费为2000元。张某某在案发后支付了李某的医疗费用15000元。\\
\textbf{Answer:}\\
根据文书内容，李某的医疗费用为15000元，后续治疗费用预计为5000元，误工费为3000元，护理费为2000元。这些费用总和为25000元。虽然张某某已经支付了15000元的医疗费用，但总犯罪金额仍为25000元，因为犯罪金额的计算是基于受害人实际遭受的损失，而非被告人已经支付的金额。$<$DTK$>$[金额]25000元$<$eoa$>$
   \end{problembox}
\end{CJK*}

\end{document}